\pdfoutput=1
\documentclass[11pt]{article}

\usepackage{acl}

\usepackage{times}
\usepackage{latexsym}
\usepackage[T1]{fontenc}
\usepackage[utf8]{inputenc}
\usepackage{microtype}

\usepackage{amsmath,amsfonts,amsthm,amssymb,bbm}
\usepackage{graphicx}
\usepackage{enumitem}
\usepackage{mathtools}
\usepackage{subfigure}
\usepackage{booktabs, multirow, bigdelim}
\usepackage{balance}
\usepackage{numprint}
\npthousandsep{,}

\title{Uncertainty Quantification with Pre-trained Language Models:\\A Large-Scale Empirical Analysis}

\author{
    Yuxin Xiao$^{1}$, Paul Pu Liang$^{2}$, Umang Bhatt$^{3}$, \\ 
    \textbf{Willie Neiswanger$^{4}$, Ruslan Salakhutdinov$^{2}$, Louis-Philippe Morency$^{2}$} \\
    $^{1}$Massachusetts Institute of Technology, $^{2}$Carnegie Mellon University, \\ $^{3}$University of Cambridge, $^{4}$Stanford University \\
    \texttt{$^{1}$yuxin102@mit.edu} \texttt{$^{2}$\{pliang,rsalakhu,morency\}@cs.cmu.edu} \\     \texttt{$^{3}$usb20@cam.ac.uk $^{4}$neiswanger@cs.stanford.edu}
}

\begin{document}
\maketitle

\begin{abstract}


Pre-trained language models (PLMs) have gained increasing popularity due to their compelling prediction performance in diverse natural language processing (NLP) tasks. 
When formulating a PLM-based prediction pipeline for NLP tasks, it is also crucial for the pipeline to minimize the \textit{calibration error}, especially in safety-critical applications. 
That is, the pipeline should reliably indicate when we can trust its predictions. 
In particular, there are various considerations behind the pipeline: (1) the choice and (2) the size of PLM, (3) the choice of uncertainty quantifier, (4) the choice of fine-tuning loss, and many more. 
Although prior work has looked into some of these considerations, they usually draw conclusions based on a limited scope of empirical studies. 
There still lacks a holistic analysis on how to compose a well-calibrated PLM-based prediction pipeline. 
To fill this void, we compare a wide range of popular options for each consideration based on three prevalent NLP classification tasks and the setting of domain shift. 
In response, we recommend the following: (1) use ELECTRA for PLM encoding, (2) use larger PLMs if possible, (3) use Temp Scaling as the uncertainty quantifier, and (4) use Focal Loss for fine-tuning.

\end{abstract}
\section{Introduction}
\label{sec:1}


PLMs \cite{qiu2020pre, min2021recent} have achieved state-of-the-art performance on a broad spectrum of NLP benchmarks \cite{rajpurkar2016squad, rajpurkar2018know, wang2019superglue, wang2019glue} and are increasingly popular in various downstream applications such as question answering \cite{yoon2019pre, garg2020tanda}, text classification \cite{arslan2021comparison, limsopatham2021effectively}, and relation extraction \cite{zhou2021atlop, xiao2021sais}.
Consequently, it is paramount for PLMs to faithfully communicate when to (or not to) rely on their predictions for decision-making, especially in high-stakes scenarios.
In these cases, we need PLMs to quantify their uncertainty accurately and calibrate well \cite{abdar2021review}, meaning that their predictive confidence should be a valid estimate of how likely they are to make a correct prediction.
Consider an example of medical question answering \cite{yoon2019pre, zhang2021smedbert} where a PLM is asked to assist doctors when diagnosing diseases.
If the PLM is $90\%$ sure that a patient is healthy, the predicted outcome should occur $90\%$ of the time in practice.
Otherwise, it may adversely affect doctors' judgment and lead to catastrophic consequences.
Hence, since PLMs have become the de facto paradigm for many NLP tasks, it is necessary to assess their calibration quality.

When constructing a well-calibrated PLM-based prediction pipeline for NLP tasks, various considerations are involved. To name a few:
\begin{enumerate}[noitemsep,topsep=0pt]
    \item Due to the use of diverse pre-training datasets and strategies, different PLMs may behave differently regarding calibration.
    \item The model size of PLMs may also affect their capability in calibration.
    \item Leveraging uncertainty quantifiers (e.g., Temp Scaling \cite{guo2017calibration} and MC Dropout \cite{gal2016dropout}) alongside PLMs in the pipeline may reduce calibration error.
    \item Some losses (e.g., Focal Loss \cite{mukhoti2020calibrating} and Label Smoothing \cite{muller2019does}) may fine-tune PLMs to calibrate better.
\end{enumerate}
Although some of these considerations have been studied before, the ideal choice for each consideration remains obscure.
On the one hand, \citet{desai2020calibration} report unconventional calibration behavior for PLMs, which casts doubts on the prior beliefs drawn on traditional neural networks by \citet{guo2017calibration}.
On the other hand, existing work \cite{desai2020calibration, dan2021effects} on PLMs' empirical calibration performance often looks at a single consideration and concludes by comparing only one or two types of PLMs.

Therefore, in this paper, we present a comprehensive analysis of the four pivotal considerations introduced above via large-scale empirical evaluations.
To ensure that our analysis is applicable to various NLP tasks and resilient to domain shift, we set up three NLP tasks (i.e., Sentiment Analysis, Natural Language Inference, and Commonsense Reasoning) and prepare both in-domain and out-of-domain testing sets for each task.
In addition to the explicit metrics of prediction and calibration error, we also utilize two evaluation tasks to examine calibration qualities implicitly.
Selective prediction lowers prediction error by avoiding uncertain testing points, and out-of-domain detection checks if a pipeline is less confident on unseen domains.
By comparing four to five options for each consideration, we recommend the following:
\begin{enumerate}[noitemsep,topsep=0pt]
    \item Use ELECTRA \cite{clark2020electra} as the PLM to encode input text sequences.
    \item Use the larger version of a PLM if possible.
    \item Use Temp Scaling \cite{guo2017calibration} for post hoc uncertainty recalibration.
    \item Use Focal Loss \cite{mukhoti2020calibrating} during the fine-tuning stage.
\end{enumerate}
Compared to prior work, our extensive empirical evaluations also reveal the following novel observations that are unique to PLM-based pipelines:
\begin{itemize}[noitemsep,topsep=0pt]
    \item The calibration quality of PLMs is relatively consistent across tasks and domains, except XLNet \cite{yang2019xlnet} being the most vulnerable to domain shift.
    \item In contrast to other NLP tasks, larger PLMs are better calibrated in-domain in Commonsense Reasoning.
    \item Uncertainty quantifiers (e.g., Temp Scaling) are generally more effective in improving calibration out-of-domain.
    \item Ensemble \cite{lakshminarayanan2017simple} is less effective in PLM-based pipelines.
\end{itemize}
To encourage future work towards better uncertainty quantification in NLP, we release our code and large-scale evaluation benchmarks containing $120$ PLM-based pipelines based on four metrics (prediction and calibration error, selective prediction, and out-of-domain detection).
These pipelines consist of distinct choices concerning the four considerations and are tested on all three NLP tasks under both in- and out-of-domain settings.\footnote{Our data and code are available at \url{https://github.com/xiaoyuxin1002/UQ-PLM.git}.}

\section{Background} 
\label{sec:2}


\subsection{Problem Formulation} 
\label{sec:2.1}

\textbf{Datasets.}
In this work, we focus on utilizing PLMs for NLP classification tasks. 
More specifically, consider such a task where the training set $\mathbb{D}_{\text{train}} = \{(x_i, y_i)\}_{i=1}^{N_\text{train}}$ consists of pairs of a text sequence $x_i \in \mathcal{X}_{\text{in}}$ and an associated label $y_i \in \mathcal{Y}$.
Similarly, the validation set $\mathbb{D}_{\text{val}}$ and the in-domain testing set $\mathbb{D}_{\text{in}}$ come from the same domain $\mathcal{X}_\text{in}$ and share the same label space $\mathcal{Y}$.
We also prepare an out-of-domain testing set $\mathbb{D}_{\text{out}}$, which differs from the others by coming from a distinct domain $\mathcal{X}_\text{out}$.

\textbf{PLM-based Pipeline.}
We apply a PLM $M$ to encode an input text sequence $x_i$ and feed the encoding vector to a classifier $F$, which outputs a predictive distribution $\mathbf{u}_i$ over the label space $\mathcal{Y}$ via the softmax operation.
Here, parameters in $M$ and $F$ are fine-tuned by minimizing a loss function $\ell$ on $\mathbb{D}_\text{train}$. 
It is optional to modify the distribution $\mathbf{u}_i$ post hoc by an uncertainty quantifier $Q$ to reduce calibration error. 
We define the predicted label as $\hat{y}_i = \arg\max_{j\in\{1,\dots,|\mathcal{Y}|\}} \mathbf{u}_{ij}$ with the corresponding confidence $\hat{c}_i = \mathbf{u}_{i\hat{y}_i}$.

\textbf{Calibration.}
One crucial goal of uncertainty quantification is to improve calibration.
That is, the predicted confidence should match the empirical likelihood: $P(y_i=\hat{y}_i \mid \hat{c}_i) = \hat{c}_i$.
We follow \citet{guo2017calibration} by using the expected calibration error (ECE) to assess the calibration performance.
The calculation of ECE is described in Section~\ref{sec:3.1}.
To reduce ECE, our main experimental evaluation lies in examining four considerations involved in a PLM-based pipeline: (1) the choice of PLM $M$ (Section~\ref{sec:3}), (2) the size of PLM $M$ (Section~\ref{sec:4}), (3) the choice of uncertainty quantifier $Q$ (Section~\ref{sec:5}), and (4) the choice of loss function $\ell$ (Section~\ref{sec:6}).


\subsection{Related Work}
\label{sec:2.2}

Uncertainty quantification has drawn long-lasting attention from various domains~\cite{bhatt2021uncertainty}, such as weather forecasting \cite{brier1950verification, raftery2005using}, medical practice \cite{yang2010nurses, jiang2012calibrating}, and machine translation \cite{ott2018analyzing, zhou2020uncertainty, wei2020uncertainty}.
Researchers have approached this question from both Bayesian \cite{kendall2017uncertainties, depeweg2018decomposition} and frequentist perspectives \cite{alaa2020discriminative, alaa2020frequentist}.
They have also proposed different techniques to improve uncertainty calibration for classification \cite{kong2020calibrated, krishnan2020improving} and regression \cite{kuleshov2018accurate, cui2020calibrated, chung2021beyond} tasks.
Recent work has investigated connections between uncertainty and other properties, such as model interpretability \cite{antoran2021getting, ley2022diverse}, selective prediction \cite{xin2021art, varshney2022investigating, varshney2022towards}, and out-of-domain generalization \cite{wald2021calibration, qin2021improving}.

PLMs \cite{qiu2020pre, min2021recent} have achieved state-of-the-art prediction performance on diverse NLP benchmarks \cite{rajpurkar2016squad, rajpurkar2018know, wang2019superglue, wang2019glue} and demonstrated many desired properties like stronger out-of-domain robustness \cite{hendrycks2020pretrained} and better uncertainty calibration \cite{desai2020calibration}.
They typically leverage a Transformer architecture \cite{vaswani2017attention} and are pre-trained by self-supervised learning \cite{jaiswal2021survey}.

Although \citet{guo2017calibration} report that larger models tend to calibrate worse, PLMs have been shown to produce well-calibrated uncertainty in practice \cite{desai2020calibration}, albeit for giant model sizes.
Their unusual calibration behavior puts the observations drawn on traditional neural networks \cite{ovadia2019can, mukhoti2020calibrating} or pre-trained vision models \cite{minderer2021revisiting} in doubt.
Prior work \cite{desai2020calibration, dan2021effects} on the calibration of PLMs often explores only one or two types of PLMs and ignores uncertainty quantifiers and fine-tuning losses beyond Temp Scaling and Cross Entropy, respectively.
As a result, there lacks a holistic analysis that explores the full set of these considerations in a PLM-based pipeline.
Therefore, our paper aspires to fill this void via extensive empirical studies.

\section{Which Pre-trained Language Model?}
\label{sec:3}

\subsection{Experiment Setup} 
\label{sec:3.1}


To evaluate the calibration performance of PLMs, we consider a series of NLP classification tasks:
\begin{enumerate}[noitemsep,topsep=0pt]
    \item \textbf{Sentiment Analysis} identifies the binary sentiment of a text sequence. 
    We treat the IMDb \textit{movie review} dataset \cite{maas2011learning} as in-domain and the Yelp \textit{restaurant review} dataset \cite{zhang2015character} as out-of-domain.
    \item \textbf{Natural Language Inference} predicts the relationship between a hypothesis and a premise. 
    We regard the Multi-Genre Natural Language Inference (MNLI) dataset \cite{williams2018broad} covering a range of genres of \textit{spoken and written text} as in-domain and the Stanford Natural Language Inference (SNLI) dataset \cite{bowman2015large} derived from \textit{image captions} only as out-of-domain.
    \item \textbf{Commonsense Reasoning} determines the most reasonable continuation of a sentence among four candidates. 
    We view the Situations With Adversarial Generations (SWAG) dataset \cite{zellers2018swag} as in-domain and its \textit{adversarial} variant (HellaSWAG) \cite{zellers2019hellaswag} as out-of-domain.
\end{enumerate}
For each task, we construct $\mathbb{D}_\text{train}$, $\mathbb{D}_\text{val}$, and $\mathbb{D}_\text{in}$ from the corresponding in-domain dataset, and $\mathbb{D}_\text{out}$ from the corresponding out-of-domain dataset.
The original validation set of each dataset is split in half randomly to form a held-out non-blind testing set (i.e., $\mathbb{D}_\text{in}$ or $\mathbb{D}_\text{out}$).
Table~\ref{tab:task} describes the task details.

\begin{table}[t!]
\setlength\tabcolsep{3pt}
  \centering
\resizebox{\columnwidth}{!}{
  \begin{tabular}{lccc}
    \toprule
        \multirow{2}{*}{} & \textbf{Sentiment} & \textbf{Natural Language} & \textbf{Commonsense} \\
        & \textbf{Analysis} & \textbf{Inference} & \textbf{Reasoning} \\
    \midrule
    \midrule
        $\mathcal{X}_\text{in}$ & IMDb & MNLI & SWAG \\
        $\mathcal{X}_\text{out}$ & Yelp & SNLI & HellaSWAG \\
        $|\mathcal{Y}|$ & 2 & 3 & 4 \\
        $|\mathbb{D}_\text{train}|$ & \numprint{25000} & \numprint{392702} & \numprint{73546} \\
        $|\mathbb{D}_\text{val}|$ & \numprint{12500} & \numprint{4907} & \numprint{10003} \\
        $|\mathbb{D}_\text{in}|$ & \numprint{12500} & \numprint{4908} & \numprint{10003} \\
        $|\mathbb{D}_\text{out}|$ & \numprint{19000} & \numprint{4923} & \numprint{5021} \\
    \bottomrule
  \end{tabular}
}
  \caption{In- and out-of-domain datasets, label space size, and each data split size of the three NLP tasks.}
  \label{tab:task}
\end{table}

\begin{table}[t!]
\setlength\tabcolsep{2pt}
  \centering
\resizebox{\columnwidth}{!}{
  \begin{tabular}{cccc}
    \toprule
        \textbf{Hugging Face} & \textbf{Model} & \textbf{Pre-training} & \textbf{Pre-training} \\
        \textbf{Name} & \textbf{Size} & \textbf{Corpus Size} & \textbf{Task} \\
    \midrule
    \midrule
        bert-base-cased & 109M & 16G & Masked LM, NSP \\
        xlnet-base-cased & 110M & 161G & Permuted LM \\
        electra-base-discriminator & 110M & 161G & Replacement Detection \\
        roberta-base & 125M & 161G & Dynamic Masked LM \\
        deberta-base & 140M & 85G & Dynamic Masked LM \\
    \midrule
        bert-large-cased & 335M & 16G & Masked LM, NSP \\
        xlnet-large-cased & 340M & 161G & Permuted LM \\
        electra-large-discriminator & 335M & 161G & Replacement Detection \\
        roberta-large & 335M & 161G & Dynamic Masked LM \\
        deberta-large & 350M & 85G & Dynamic Masked LM \\
    \bottomrule
  \end{tabular}
}
  \caption{Model size, pre-training corpus size, and pre-training task of the five PLMs, separated into the base (upper) and the large (lower) versions.}
  \label{tab:model}
\end{table}

\begin{figure*}[t!]
    \centering
    \includegraphics[width=0.765\textwidth]{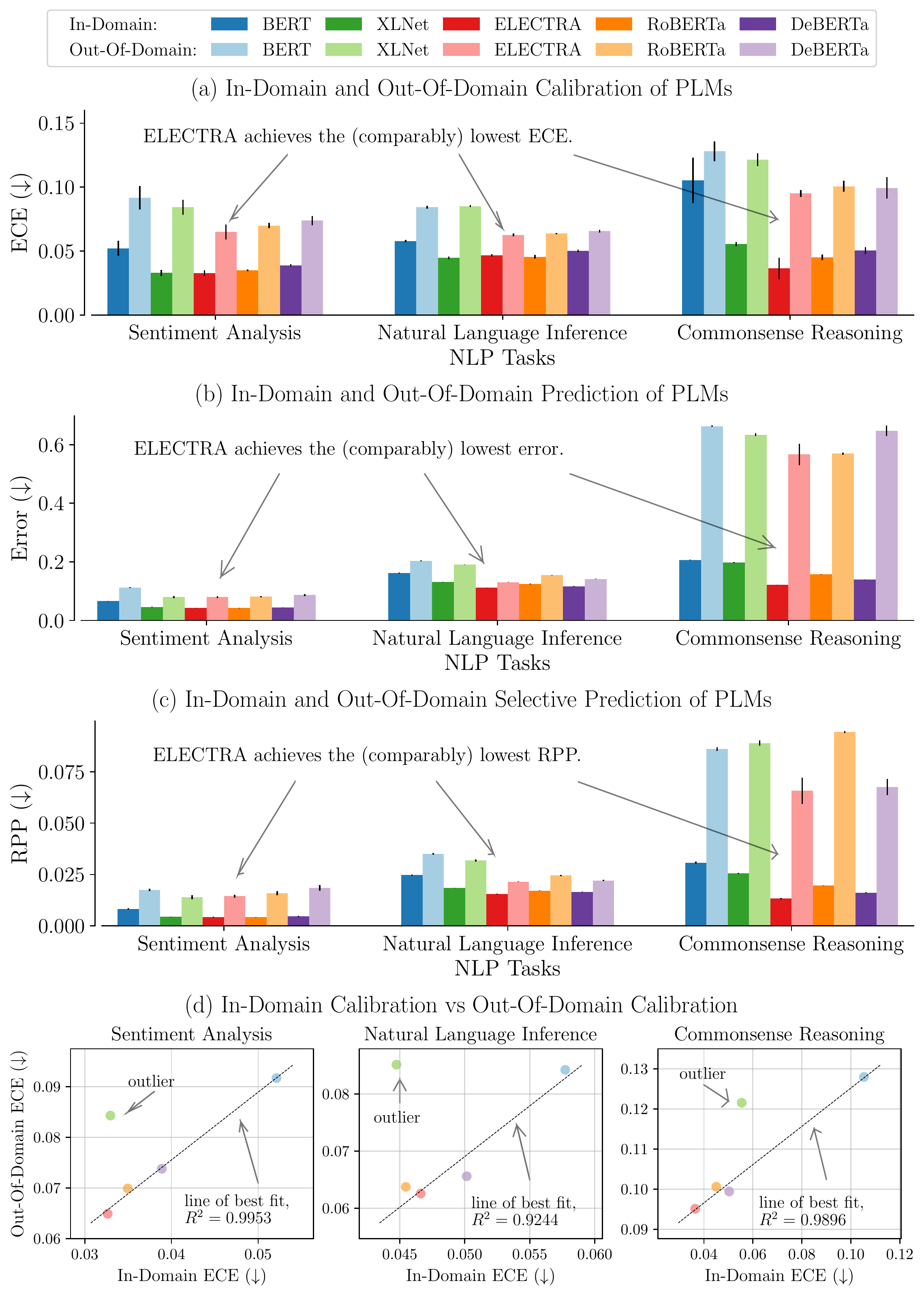} 
    \caption{Calibration and (selective) prediction performance of five PLMs in three NLP tasks under two domain settings. The calibration quality of the five PLMs is relatively consistent across tasks and domains, while XLNet is the least robust to domain shift. ELECTRA stands out due to its lowest scores in ECE, prediction error, and RPP.}
    \label{fig:model}
\end{figure*}

\begin{figure*}[t!]
    \centering
    \includegraphics[width=0.725\textwidth]{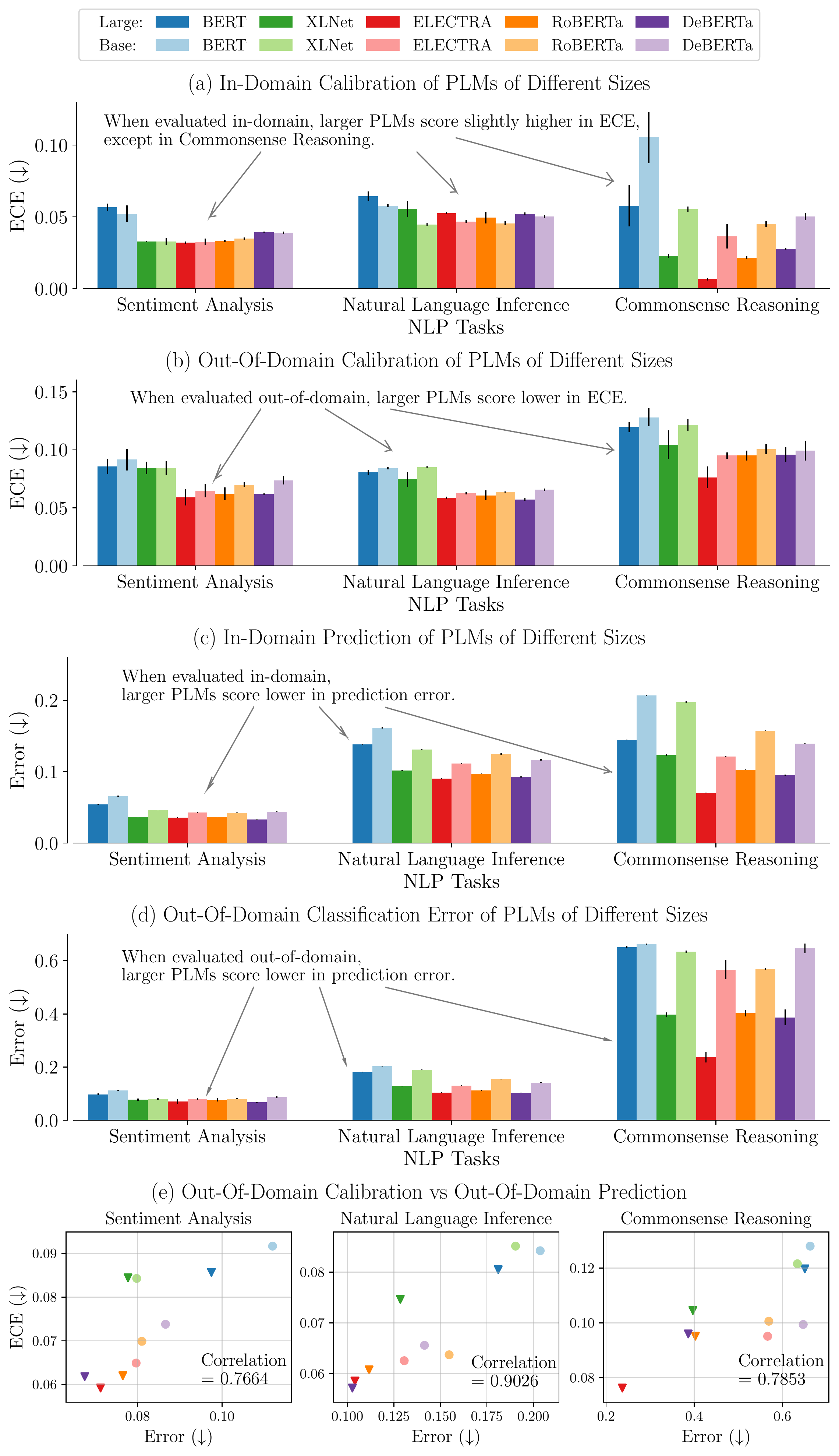} 
    \caption{Calibration and prediction performance of large and base PLMs in three NLP tasks under two domain settings. Larger PLMs calibrate better than their respective base versions when evaluated out-of-domain, while calibrating slightly worse in-domain with one exception in Commonsense Reasoning. If the computational budget permits, larger PLMs constitute more powerful pipelines given their lower out-of-domain ECE along with lower prediction error. We also observe a positive correlation between calibration and prediction error out-of-domain.}
    \label{fig:size}
\end{figure*}

To understand which PLM delivers the lowest calibration error, we examine five popular options:
\begin{enumerate}[noitemsep,topsep=0pt]
    \item \textbf{BERT} \cite{devlin2019bert} utilizes a bidirectional Transformer architecture pre-trained by masked language modeling (LM) and next sentence prediction (NSP).
    \item \textbf{XLNet} \cite{yang2019xlnet} proposes a two-stream self-attention mechanism and a pre-training objective of permuted LM.
    \item \textbf{ELECTRA} \cite{clark2020electra} pre-trains a discriminative model to detect tokens replaced by a generative model.
    \item \textbf{RoBERTa} \cite{liu2019roberta} builds on BERT by pre-training based on dynamic masked LM only and tuning key hyperparameters.
    \item \textbf{DeBERTa} \cite{he2020deberta} further improves RoBERTa via a disentangled attention mechanism and an enhanced mask decoder.
\end{enumerate}
We use the base version of each PLM, which has a similar model size and is initialized from the corresponding Hugging Face \cite{wolf2020transformers} pre-trained checkpoint.
Table~\ref{tab:model} details these PLMs.
After receiving the encoding vector of the classification token $\texttt{[CLS]}$ for an input text sequence from the PLM, we pass it through a classifier to obtain a predictive distribution. 
Regarding the classifier configuration, we follow the default practice in Hugging Face by utilizing a two-layer neural network with tanh non-linear activation.

The learning rate for each model-dataset combination is tuned based on the validation set among $\{5\mathrm{e}{-6}, 1\mathrm{e}{-5}, 2\mathrm{e}{-5}, 5\mathrm{e}{-5}\}$.
We leverage AdamW \cite{loshchilov2018decoupled} to minimize the cross-entropy loss on $\mathbb{D}_\text{train}$ for five epochs with early stopping and a linearly decaying scheduler \cite{goyal2017accurate} whose warm-up ratio $=10\%$.
Batch size is $16$, and the model gradients are clipped to a maximum norm of $1$.
We perform our experiments on a Tesla A6000 GPU and report the mean and one standard error by conducting six trials with different seeds.

To explicitly evaluate calibration performance by ECE, we first stratify $N$ predictions into $K$ bins of equal width based on the sorted confidence values.
Then ECE is a weighted average of the absolute difference between the accuracy and confidence of each bin: $\text{ECE} = \sum_{k=1}^K \frac{|B_k|}{N} |\text{acc}(B_k) - \text{conf}(B_k)|$, where $\text{acc}(B_k)$ and $\text{conf}(B_k)$ are the average accuracy and confidence of predictions in bin $B_k$, respectively.
We set $K=10$ in our experiments.

To implicitly assess calibration quality based on selective prediction, we deploy the metric of reversed pair proportion (RPP) \cite{xin2021art}.
More specifically, for a dataset of size $N$, $\text{RPP} = \frac{1}{N^2} \sum_{i=1}^N \sum_{j=1}^N \mathbbm{1}[\hat{c}_i<\hat{c}_j, y_i=\hat{y}_i, y_j\neq\hat{y}_j]$.
It measures the proportion of prediction pairs with a reversed confidence-error relationship.
A lower RPP indicates that the pipeline is more confident on correct predictions.

\subsection{Empirical Findings}
\label{sec:3.2}

As shown in Figure~\ref{fig:model}(a), the calibration performance of all five PLMs deteriorates from in-domain to out-of-domain.
This phenomenon coincides with the finding made by \citet{ovadia2019can} on traditional neural networks.
In addition, \textbf{the ranking among the five PLMs based on ECE is generally consistent,} which implies that their calibration quality is transferable across tasks and domains.
More specifically, for all three tasks under the in-domain setting, XLNet, ELECTRA, RoBERTa, and DeBERTa outperform BERT in terms of lower ECE, suggesting that a larger pre-training corpus may improve the calibration quality (see Table~\ref{tab:model}).
\textbf{When moving to the out-of-domain setting, XLNet sees the largest increase in ECE,} which makes it an outlier in Figure~\ref{fig:model}(d).
This observation may indicate that the pre-training task of permuted LM is vulnerable to domain shift.

\textbf{ELECTRA stands out among the five examined PLMs in encoding input text sequences.}
Not only does it achieve the (comparably) lowest ECE in all three tasks under both in- and out-of-domain settings, it also delivers the lowest prediction error in Figure~\ref{fig:model}(b) and the lowest RPP for selective prediction in Figure~\ref{fig:model}(c).
We hypothesize its success to the unique pre-training paradigm of replaced token detection, which preserves the token distribution by avoiding the artificial $\texttt{[MASK]}$ tokens in masked LM and enhances the computational efficiency by learning from all input tokens.

\section{What Model Size?}
\label{sec:4}

\subsection{Experiment Setup}
\label{sec:4.1}


To investigate how the size of PLMs affects the calibration performance, we compare the large versions of the five PLMs mentioned in Section~\ref{sec:3.1} against their respective base versions.
We keep the rest of the setup the same as in Section~\ref{sec:3.1}.

\subsection{Empirical Findings}
\label{sec:4.2}


Figures~\ref{fig:size}(a) and (b) demonstrate that larger PLMs tend to produce a slightly higher ECE compared to their respective base versions when evaluated in-domain, while calibrating better out-of-domain.
This observation based on five PLMs verifies the conclusion made by \citet{dan2021effects} solely based on BERT.
However, \textbf{there is a notable exception that larger PLMs are significantly better calibrated in-domain in Commonsense Reasoning than their respective base versions,} which implies that larger PLMs are more aware of their uncertainties during the reasoning process.

\textbf{Larger PLMs constitute more powerful PLM-based pipelines, if computational budget permits.}
Although sometimes they suffer slightly in in-domain calibration compared to their smaller counterparts, larger PLMs achieve a lower ECE out-of-domain.
They also deliver lower in- and out-of-domain prediction errors in Figures~\ref{fig:size}(c) and (d), respectively.
In addition, we observe a positive correlation between calibration and prediction errors under the out-of-domain setting in Figure~\ref{fig:size}(e), suggesting that pipelines calibrating well out-of-domain are more accurate under domain shift as well.
This reflects the finding in \citet{wald2021calibration} that multi-domain calibration leads to better out-of-domain prediction performance.
\begin{figure*}[t!]
    \centering
    \includegraphics[width=0.785\textwidth]{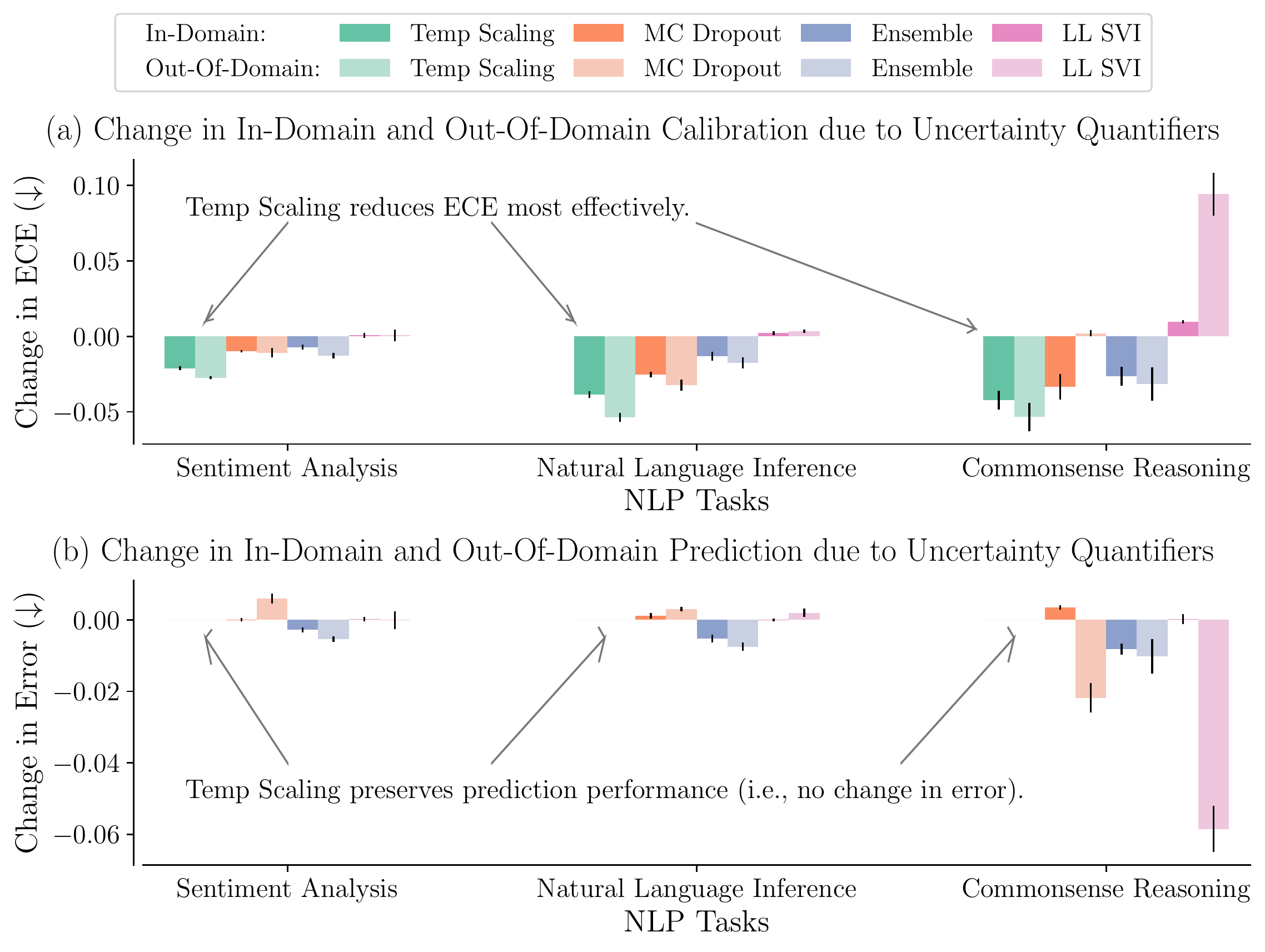} 
    \caption{Change in calibration and prediction performance due to the use of four uncertainty quantifiers. The effectiveness of these quantifiers in reducing ECE follows the descending order of Temp Scaling, MC Dropout, Ensemble, and LL SVI. The drop in ECE is more significant out-of-domain. Temp Scaling is the most compelling fine-tuning loss due to its largest reduction in ECE, preservation of prediction results, and little computational cost.}
    \label{fig:quantifier}
\end{figure*}

\section{Which Uncertainty Quantifier?}
\label{sec:5}

\subsection{Experiment Setup} 
\label{sec:5.1}


As discussed in Section~\ref{sec:2.1}, we can further adjust the vanilla predictive distribution post hoc via an uncertainty quantifier.
Therefore, we study four uncertainty quantifiers based on the setup in Section~\ref{sec:3.1} to inspect which improve the calibration performance in our problem formulation:
\begin{enumerate}[noitemsep,topsep=0pt]
    \item \textbf{Temp Scaling} \cite{guo2017calibration} learns a scalar parameter $T_\text{temp}$ based on $\mathbb{D}_\text{val}$ and ``softens'' the vanilla logit output with $T_\text{temp}$ to obtain a new predictive distribution.
    \item \textbf{MC Dropout} \cite{gal2016dropout} approximates the expectation of a posterior predictive distribution by averaging $T_\text{mc}$ forward passes with dropout turned on.
    \item \textbf{Ensemble} \cite{lakshminarayanan2017simple} averages the predictive distributions of $T_\text{en}$ independently trained models.
    \item \textbf{LL SVI} (Last-Layer Stochastic Variational Inference) \cite{blundell2015weight} implements variational layers with reparameterized Monte Carlo estimators based on the Bayesian-Torch package \cite{krishnan2022bayesiantorch}. It approximates the expectation of a posterior predictive distribution by averaging $T_\text{svi}$ forward passes through the Bayesian classification layers. 
\end{enumerate}
Here, we follow \citet{lakshminarayanan2017simple} by setting $T_\text{en} = 5$. 
We use $T_\text{mc} = 10$ and $T_\text{svi} = 50$ due to computational constraints during inference.
The dropout rate in MC Dropout is the same as the default dropout rate of each PLM.

\begin{figure*}[t!]
    \centering
    \includegraphics[width=0.785\textwidth]{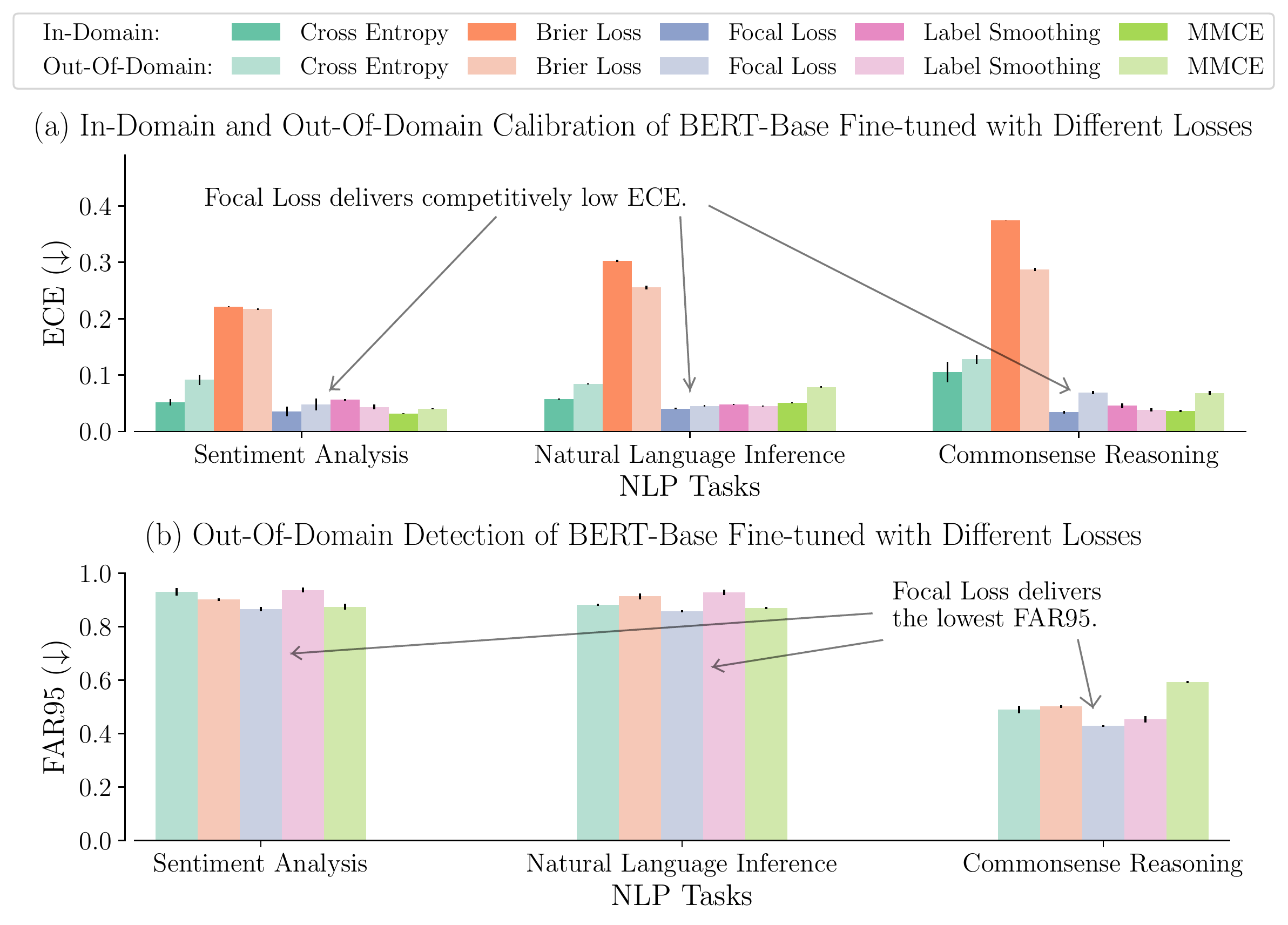}
    \caption{Calibration and out-of-domain detection performance of BERT base models fine-tuned by five losses. Focal Loss, Label Smoothing, and MMCE are more capable of fine-tuning well-calibrated models compared to Cross Entropy and Brier Loss. Focal Loss is the best option due to its competitively low ECE and FAR95.}
    \label{fig:loss}
\end{figure*}

\subsection{Empirical Findings}
\label{sec:5.2}

In Figure~\ref{fig:quantifier}, we plot the change in calibration and prediction performance due to the use of uncertainty quantifiers compared to the vanilla results in Section~\ref{sec:4.1}.
\textbf{The improvement in calibration is more significant out-of-domain.}
More specifically, \textbf{the degree to which these quantifiers decrease ECE follows the descending order of Temp Scaling, MC Dropout, Ensemble, and LL SVI.}
In fact, LL SVI even hurts the calibration in terms of an increase in ECE, suggesting that variational classifiers with reparameterized Monte Carlo estimators cannot capture uncertainties well when used only at the fine-tuning stage.
Unlike \citet{ovadia2019can}, \textbf{we find Ensemble less effective in PLM-based pipelines,} possibly because individual learners in Ensemble are initialized from the same pre-trained model checkpoint and, consequently, the strong correlation among them limits the power of Ensemble \cite{liu1999ensemble}.

Meanwhile, Temp Scaling preserves prediction results, and Ensemble lowers prediction error, as expected.
Although MC Dropout and LL SVI reduce the prediction error out-of-domain in Commonsense Reasoning by producing sharper predictive distributions, they usually end up being overconfident, which leads to the rise in ECE in Figure~\ref{fig:quantifier}(a).

\textbf{Temp Scaling is the most appropriate uncertainty quantifier for PLM-based pipelines.}
Compared to LL SVI, Temp Scaling diminishes ECE and maintains the competitive prediction quality of PLMs.
Moreover, the post hoc recalibration manner of Temp Scaling adds little to the computational burden.
In contrast, Ensemble or MC Dropout significantly increases the computational cost during fine-tuning or inference, respectively.
Note that this distinction is of great importance given the enormous computational burdens of PLMs.

\section{Which Fine-tuning Loss?}
\label{sec:6}

\subsection{Experiment Setup}
\label{sec:6.1}


Besides cross-entropy loss, we consider four other losses when fine-tuning a BERT base model and compare their calibration performance based on the setup in Section~\ref{sec:3.1}.
\begin{enumerate}[noitemsep,topsep=0pt]
    \item \textbf{Cross Entropy} \cite{good1952rational} is the negative log likelihood of ground-truth classes.
    \item \textbf{Brier Loss} \cite{brier1950verification} is the squared difference between predictive distributions and one-hot ground-truth vectors.
    \item \textbf{Focal Loss} \cite{mukhoti2020calibrating} applies a modulating term to cross-entropy loss to focus model learning on hard misclassified samples.
    \item \textbf{Label Smoothing} \cite{muller2019does} produces targeting distributions by allocating probability mass to non-ground-truth classes.
    \item \textbf{MMCE} (Maximum Mean Calibration Error) \cite{kumar2018trainable} is a differentiable proxy to regularize calibration error, usually used alongside cross-entropy loss.
\end{enumerate}
We use a smoothing factor of $0.1$, and follow the practice in \citet{mukhoti2020calibrating} by setting the focal hyperparameter to $5$ when the predictive probability for the ground-truth class $\in [0,0.2)$ and to $3$ when the probability $\in [0.2,1]$.

In addition, we leverage out-of-domain detection to implicitly examine the quality of uncertainty quantification.
We want models to be less confident on $\mathbb{D}_\text{out}$ than on $\mathbb{D}_\text{in}$ and, hence, report the false alarm rate at $95\%$ recall (FAR95) \cite{hendrycks2020pretrained}.
This metric tells the ratio of samples in $\mathbb{D}_\text{in}$ whose confidence is lower than the $95$th percentile of samples in $\mathbb{D}_\text{out}$.

\subsection{Empirical Findings}
\label{sec:6.2}

As shown in Figure~\ref{fig:loss}(a), \textbf{Label Smoothing, Focal Loss, and MMCE generate better-calibrated BERT base models compared to Cross Entropy and Brier Loss.}
While models fine-tuned by Cross Entropy, Focal Loss, or MMCE calibrate better in-domain, Brier Loss and Label Smoothing enjoy a decrease in ECE when evaluated out-of-domain.
This observation matches the findings in~\citet{desai2020calibration, dan2021effects} and is intuitive for Label Smoothing since it deliberately alleviates overconfidence during fine-tuning.

\textbf{Focal Loss is the most compelling fine-tuning loss for PLM-based pipelines.}
Among the five examined options, Focal Loss delivers competitively low ECE, both in- and out-of-domain for all three tasks.
Moreover, it scores the lowest in FAR95, as illustrated in Figure~\ref{fig:loss}(b), meaning that models fine-tuned by Focal Loss are most alert to domain shift.
We note that FAR95 scores are relatively high in Sentiment Analysis and Natural Language Inference, probably because these pipelines also predict well out-of-domain in Figure~\ref{fig:size}(d).
\section{Conclusion}
\label{sec:7}


In this paper, we contribute a comprehensive analysis on how to reduce calibration error in a PLM-based pipeline.
We establish four key considerations behind the pipeline and compare a broad range of prevalent options for each consideration.
Our empirical evaluations consist of three distinct NLP classification tasks and two different domain settings.
Based on our large-scale systematic analysis, we recommend the following:
\begin{enumerate}[noitemsep,topsep=0pt]
    \item Use ELECTRA for PLM encoding.
    \item Use larger PLMs if possible.
    \item Use Temp Scaling for post hoc recalibration.
    \item Use Focal Loss during the fine-tuning stage.
\end{enumerate}
Compared to existing work, we also observe the following novel phenomena that are unique to PLM-based pipelines:
\begin{itemize}[noitemsep,topsep=0pt]
    \item The relative calibration quality of PLMs is consistent in general across tasks and domains, with an exception of XLNet, which is the least robust to domain shift.
    \item Larger PLMs are better calibrated under the in-domain setting in Commonsense Reasoning, unlike in the other NLP tasks.
    \item Uncertainty quantifiers are generally more effective in improving calibration performance under the out-of-domain setting.
    \item Ensemble is less effective in reducing calibration error when used with PLM-based pipelines, despite their convincing performance with traditional models.
\end{itemize}

\section{Limitation}
\label{sec:8}

Due to computational constraints, we are unable to pre-train PLMs from scratch with other combinations of pre-training corpora and tasks.
Consequently, while our analysis is applicable to existing widely-used PLMs, we do not claim its generalization to new combinations of pre-training corpora and tasks. 
We believe that this does not invalidate our claims which are primarily targeted toward real-world practitioners using existing PLMs.
It is possible that techniques catering to the special needs of PLM-based pipelines \cite{kong2020calibrated} can mitigate calibration error further.

Moreover, although our setup involves domain shift, we do not focus on inspecting how the degree of domain shift affects the calibration performance of PLM-based pipelines.
It is also interesting to consider how to construct a well-calibrated PLM-based pipeline for other types of NLP tasks such as cross-lingual text classification and generation, which we leave to future work.

\section*{Acknowledgements}
This material is based upon work partially supported by National Science Foundation (Awards \#1722822 and \#1750439) and National Institutes of Health (Awards \#R01MH125740, \#R01MH096951, and \#U01MH116925). PPL is partially supported by a Facebook PhD Fellowship and a Carnegie Mellon University's Center for Machine Learning and Health Fellowship. Any opinions, findings, conclusions, or recommendations expressed in this material are those of the author(s) and do not necessarily reflect the views of the sponsors, and no official endorsement should be inferred.

\bibliographystyle{acl_natbib}
\bibliography{main}

\newpage
\appendix
\section{Responsible NLP Research} 
\label{app:1}

In this paper, we aim to identify the best choice for each consideration in constructing a well-calibrated PLM-based pipeline via extensive empirical studies.
Our empirical analysis involves training multiple large-scale PLMs and, consequently, consumes a fair amount of computational power.
However, we believe that the takeaways from our analysis will benefit NLP practitioners at large, which will write off the computational cost in the future.

In particular, the Hugging Face package leveraged in our experiments utilizes the Apache License 2.0, and the Bayesian-Torch package utilizes the BSD 3-Clause License.
We focus on PLM-based pipelines targeting English and assess them based on six NLP datasets, which aligns with the intended use of these datasets.
We also release the evaluation benchmarks of our empirical analysis to illustrate the performance of different PLM-based pipelines based on diverse metrics.
The benchmarks do not contain information that uniquely identifies individual people or offensive content.

\end{document}